\pdfoutput=1

\documentclass[11pt]{article}

\usepackage{acl}

\usepackage{times}
\usepackage{latexsym}

\usepackage[T1]{fontenc}

\usepackage[utf8]{inputenc}

\usepackage{microtype}

\usepackage{inconsolata}

\usepackage{microtype}
\usepackage{xcolor}
\usepackage{amsmath}
\usepackage{multirow}
\usepackage{booktabs}
\usepackage{graphicx}
\usepackage{makecell}
\usepackage{algorithm}
\usepackage{algpseudocode}
\usepackage{subcaption}
\usepackage{authblk}

\usepackage{url}

\usepackage{breakurl}
\usepackage[breaklinks]{hyperref}

\title{Multi-Step Dialogue Workflow Action Prediction}

\author[ ]{\textbf{Ramya Ramakrishnan} \quad \textbf{Ethan R. Elenberg} \quad \textbf{Hashan Narangodage} \quad \textbf{Ryan McDonald}}
\affil[ ]{ASAPP}
\affil[ ]{\texttt{\{rramakrishnan,eelenberg,hnarangodage,rmcdonald\}@asapp.com}}

\begin{document}
\maketitle
\begin{abstract}

In task-oriented dialogue, a system often needs to follow a sequence of actions, called a \textit{workflow}, that complies with a set of guidelines in order to complete a task. In this paper, we propose the novel problem of multi-step workflow action prediction, in which the system predicts multiple future workflow actions. Accurate prediction of multiple steps allows for multi-turn automation, which can free up time to focus on more complex tasks. We propose three modeling approaches that are simple to implement yet lead to more action automation: 1) fine-tuning on a training dataset, 2) few-shot in-context learning leveraging retrieval and large language model prompting, and 3) zero-shot graph traversal, which aggregates historical action sequences into a graph for prediction. We show that multi-step action prediction produces features that improve accuracy on downstream dialogue tasks like predicting task success, and can increase automation of steps by 20\% without requiring as much feedback from a human overseeing the system.

\end{abstract}

\section{Introduction}

\begin{figure}[t]
    \centering
    \includegraphics[width=0.45\textwidth]{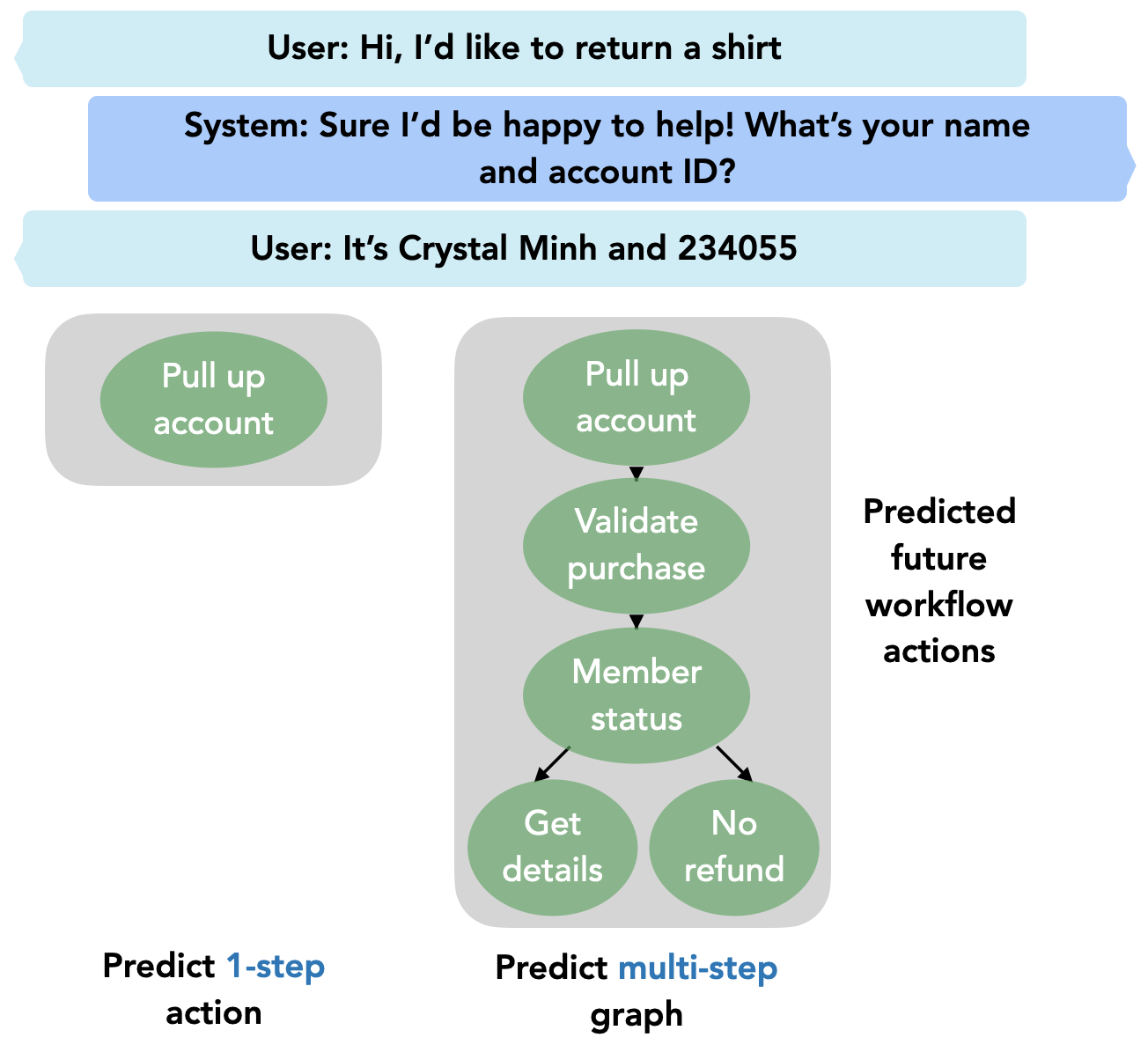}
    \caption{We propose the problem of multi-step Action State Tracking (AST), which involves predicting many future workflow actions while prior work only predicts one step. We represent predictions as graphs that capture potential branching in future action sequences.}
    \label{fig:intro_fig}
\end{figure}

Task-oriented dialogue involves understanding a user's intent, determining the right actions to take, and responding appropriately back to the user. To solve the task correctly, the system (either human or machine) needs to accurately identify the user's problem and follow a set of guidelines to ensure it's resolved appropriately. This implicit sequence of system actions, which needs to comply with business policies and guidelines, is called a \textit{workflow} \cite{hattami2022workflow}. For example, in the customer service use case shown in Figure \ref{fig:intro_fig}, a customer may ask to return a shirt, which would require agents to follow this workflow: (1) pull up account, (2) validate purchase, (3) check customer's membership status, (4) get details to offer a refund if possible.

Prior work \cite{chen2021action,hattami2022workflow} has formalized Action State Tracking (AST), which involves predicting a \textit{single} workflow action, represented as an action along with respective slot values given the past conversation context and actions. In this paper, we formalize the problem of \textit{multi-step} AST, which involves predicting the next several steps that the system must take towards task completion (\emph{e.g.} pull up account, validate purchase, check membership). This increases automation opportunities, which can make a huge impact in applications like customer service and healthcare by increasing efficiency of workers’ time, enabling human agents to focus on more complex tasks, and dramatically reducing business costs.\footnote{Labor costs can range anywhere from \$0.6 to \$0.75 per minute or \$36 to \$45 per hour \cite{fcbcoManagingCustomer,xu2020ai}.} Automation of actions and workflows can also lead to better estimates of conversational statistics, such as probability of successful task completion, allowing for optimizations such as issue escalation prediction or global scheduling of issues in a contact center.

Multi-step workflow prediction is challenging for several reasons. First, we need to predict multiple system actions without observing intermediate customer utterances. This means we have to deal with future uncertainty stemming from the customer steering the conversation in many ways. Applying single-step AST in this case would not work well as it cannot handle this uncertainty and simulate many possible futures. In this work, we propose using graphs as a way to capture branching in future action sequences, as shown in Figure~\ref{fig:intro_fig}. Another difference with prior work is that we predict system actions rather than utterances, as is commonly done with large language models \cite{zhao2020knowledge,hosseini2020simple}. Predicting system actions is relatively under-explored \cite{chen2021action,hattami2022workflow}. For this problem, we need to build models that explicitly learn the structure in workflow action sequences.

We investigate three simple modeling approaches for solving multi-step AST. The first involves fine-tuning a T5 model to predict all workflow actions until the end of the dialogue given the conversation history. The second is a few-shot in-context learning approach, which retrieves similar examples and prompts a large language model to predict the rest of the action sequence. The final is a zero-shot approach that involves offline graph construction with graph traversal at inference time. Since multi-step AST is a more challenging problem, we propose new evaluation metrics that lead to more fine-grained comparisons and measure how well our approaches capture future action sequence uncertainty. We show that our proposed methods outperform prior work on previous and new metrics for two datasets: ABCD \cite{chen2021action} and MultiWoz \cite{budzianowski2018multiwoz}.

Summarizing our contributions, we (1) propose and formalize a new problem, multi-step AST, (2) present new evaluation metrics that capture uncertainty in multi-step workflows, (3) discuss three modeling approaches, and (4) conduct extensive experiments to show that these simple modeling changes can achieve up to 20\% additional automation on system actions and utterances. Data and code will be released upon publication.

\section{Related work}

\paragraph{Dialogue state tracking:} Dialogue state tracking (DST) \cite{rastogi2017scalable,ren2018towards,nouri2018toward,mrkvsic2016neural,rastogi2017scalable} is a key component of task-oriented dialogue systems and involves identifying the user's goal or intent throughout a dialogue. Models trained on DST tasks require learning user intents (\textit{e.g.} booking a restaurant) and also extracting slot values from the user (\textit{e.g.} \texttt{reservation-time=18:00}). Traditional DST \cite{stolcke2000dialogue,henderson2014word,henderson2013deep} has been represented as an isolated task with only one speaker's state being tracked in the conversation and often requires dialogue states to be known in advance. Recent approaches \cite{bordes2016learning} involve more implicit methods that bypass DST through end-to-end approaches, for example that leverage seq2seq models. While DST is similar to our problem of predicting workflow actions and slot-values, Action State Tracking \cite{chen2021action,hattami2022workflow}, discussed next, requires leveraging user intents and agent guidelines in order to predict agent workflow actions.

\paragraph{Action state tracking:} Prior work \cite{chen2021action}  has proposed the problem of Action State Tracking (AST), which involves predicting the next action the agent should take based on the conversation and agent guidelines. 
The authors propose a few approaches to AST, including one based on RoBERTa~\cite{liu2021roberta}.
Another work \cite{hattami2022workflow} also addresses the AST problem and demonstrates improved performance with a T5 model. Both of these works focus on 1-step prediction rather than multi-step prediction. We show that with our multi-step models, we are able to achieve similar performance to the baselines on 1-step prediction, while also predicting the full sequence more accurately. We also show that multi-step prediction can increase automation of system steps by 20\%, as compared to 1-step prediction.

\paragraph{Dialogue response generation:} Response generation \cite{zhang2019dialogpt,lian2019learning,cai2019retrieval,cao2020pretrained} is an important problem in task-oriented dialogue as well. Many works \cite{chen2022dialogved,zhou2017mechanism} focus on generating relevant and diverse responses given a dialogue context. Controlled response generation \cite{wu2021controllable,ramakrishnan2022long,gupta2020controlling} goes further to control responses towards defined styles or lexical constraints like inclusion of specific words or phrases. These prior works have used similar seq2seq models for response generation, which we apply to the related problem of multi-step action prediction.

\section{Problem formulation}\label{sec:formulation}

We formalize the problem of multi-step Action State Tracking (multi-step AST) for dialogue. Given a dialogue context $C_t = [x_1,x_2,...x_t]$, our goal is to predict all future actions that will occur during the rest of the conversation $[a_{t+1},...,a_{t+N}]$. Here, each item in the context $x_i$ can either be a past system utterance $s_i$, a past user utterance $u_i$, or a past action $a_i$. With slight abuse of notation, each action $a = (a, \mathbf{v})$ includes an action slot $a$ (\emph{e.g.} \texttt{pull-up-account}), and an optional list of action values $\mathbf{v}$ (\emph{e.g.} \texttt{[crystal minh]}). 

To predict a multi-step sequence of future actions, we assume access to a historical dataset of dialogues paired with ground truth action sequences. We also assume that dialogues may be grouped according to the customer's intent, which helps determine relevant actions for resolving the issue. Given the context up to step $t$, a multi-step prediction is a weighted, directed graph with vertices $V=\{a_{t+1},...,a_{t+N}\}$ and edges $E=\{e_{ij}\}$ representing the probability that action $a_i$ is followed by action $a_j$. Every path in the graph corresponds to a possible chain of actions, and branches represent uncertainty in how the customer's subsequent utterances will influence the outcome of the conversation. Prior work primarily focuses on generating a single chain of future actions $[a_{t+1},...,a_{t+N}]$, which is a special case of our graph output.
 
Being able to predict graphs for multi-step AST is important since a single chain captures only one likely future path. It does not capture possible branching in the future action sequence occurring from customer utterances (\emph{e.g.} gold vs. silver member) or from external knowledge of guidelines (\emph{e.g.} refunds are only valid for 60 days after the purchase date). In contrast, graphs can represent this inherent uncertainty in dialogue by aggregating many possible chains together.

\section{Multi-step AST Approaches}

Next, we describe each modeling approach. All approaches use a dataset of (conversation, workflow) pairs and returns a directed graph. 

\subsection{Fine-tuning} \label{sec:fine_tuning}
The first approach involves fine-tuning a language model to predict all future workflow actions. We construct training examples of the form (context, action sequence) and train a sequence to sequence model with the following inputs:\vspace{-0.2cm}
\begin{align*}
\text{Source: } x_1, x_2, ..., x_t \;\; .
\end{align*}
Following \citet{hattami2022workflow}, we use the source prefix \texttt{Predict AST:} and the following target:\vspace{-0.1cm}
\begin{align*}
    \text{Target: } &  a_{t+1}[v_{t+1}^1, ..., v_{t+1}^{m_{t+1}}]; \ldots \\
    &\qquad \ldots ; \, a_{t+N}[v_{t+N}^1, ..., v_{t+N}^{m_{t+N}}] \;\; ,
\end{align*}
where $a_t$ refers to the action type, $v_t^i$ includes all $m_t$ values for each action value. In this equation, $a_{t+1}$ refers to the action name for timestep $t+1$ (e.g., verify-identity, pull-up-account) and $a_{t+N}$ refers to the action name for timestep $t+N$. Each action has a variable number of slot values, and the upper index $m_{t+1}$ is the number of slot values for the corresponding action at timestep $t+1$ (e.g., verify-identity(full-name, account-id, order-id) has 3 slot values) and similarly for $m_{t+N}$. The input is the full dialogue context until the current time step and the output is the rest of the actions in the dialogue. We then fine-tune a model to predict all future workflow actions. This generalizes prior work~\cite{hattami2022workflow,chen2021action} by considering cases where $N > 1$.

At inference time, we generate $R$ model rollouts. The graph is constructed by counting all transitions in the rollouts as edges (Algorithm 1). The graph root node is the last executed action in the context. For the 1-step AST model, we use the same training procedure, but every sequence is of size one so it cannot capture more than the next action.

\subsection{Few-shot in-context learning}
This approach does not require fine-tuning, and instead relies entirely on predictions from large pretrained language models. This includes cloud-based APIs for which model weights are not publicly available. First, we set a text representation for all examples in the dataset, \emph{i.e.} utterance and/or action contexts. Then, we index all training examples by embedding this representation with a pretrained \emph{retrieval} model such as SBERT~\cite{reimers2019sbert}. At inference time, we retrieve the top $K$ training examples $(\text{Source}_1, \text{Target}_1), \ldots (\text{Source}_K, \text{Target}_K)$ that have highest cosine similarity with the test example. The utterances and/or actions from the retrieved examples are concatenated with the test input, along with an instruction to predict future workflow actions of the dialogue. We pass the following prompt to a pretrained \emph{prediction} model:

{
\scriptsize
\sf{``You are a helping a user with a customer service issue. Predict the sequence of actions the system should take in the future. Follow the format in the examples: add optional values in square brackets and add a semicolon between actions.{\textbackslash n}{\textbackslash n}Example\#1:{\textbackslash n}\{\{example1\}\}...
}}

where each example is one of the retrieved top $K$ from the training set. The directed subgraph is constructed in a similar way to the fine-tuning method where $R$ rollouts are generated and combined into a subgraph of future possible action sequences.

\begin{algorithm}[t!]
\caption{Graph construction algorithm}\label{alg:graph_traversal}
\begin{algorithmic}
\Require $\mathcal{P} = [p_1,..., p_{n_p}]$ \Comment{Policies}
\Require $\{\mathcal{W}_{p}\}$ \Comment{Workflows}
\Require $t_{edge}$ \Comment{Edge threshold}
\For{policy $p$ in $\mathcal{P}$}

\State $n_{edge}$ = \{\} \Comment Edge counts
\For{workflow $w$ in $\mathcal{W}_p$}
\For{action $a_t$ in $w$=[$a_0$,...,$a_T$]}
\State $n_{edge}[(a_t, a_{t+1})]$\ += 1
\EndFor
\EndFor
\State graph\_edges = \{$e$ | $n_{edge}[e] \geq t_{edge}$\}
\EndFor
\end{algorithmic}
\end{algorithm}

\subsection{Zero-shot graph traversal}

Rather than using a language model, this approach groups historical data by customer intent or policy and pre-computes a graph of relevant actions for each policy $p \in \mathcal{P}$. This is a reasonable assumption because in many customer service applications, the intent is often known after the first customer utterance and used for routing to the right human agent. The graph captures possible future workflow action paths depending on previously completed actions.

Algorithm \ref{alg:graph_traversal} includes the pseudocode for the approach. For each policy, we loop through all workflows in that policy and compute counts of action transitions. We keep all edges that appear more than $t_{edge}$ times. At inference time, we traverse the graph using the past workflow actions in the context. The final graph prediction is the graph that starts at the last workflow action and includes all future edges appearing more than $t_{edge}$ times. This approach will deterministically return the same most likely graph given the same past action sequence. Note that utterance contexts are never used explicitly, only implicitly via previous actions. This assumes that workflows are regular and can be predicted well purely based on what actions have occurred so far. Since graphs can include cycles, we set the max predicted sequence length based on the action sequence lengths observed in the training data. This approach does not require rollouts and is completely zero-shot, as it requires neither training examples nor prompting.

\section{Experiments}
\label{sec:experiments}
\subsection{Datasets}
We run experiments on two task-oriented dialogue datasets: ABCD \cite{chen2021action} and MultiWoz \cite{budzianowski2018multiwoz}. Both datasets have dialogues along with corresponding workflows constructed by \newcite{hattami2022workflow}. We use these workflows as ground truth and train models to do multi-step Action State Tracking.\vspace{-0.3cm}
\paragraph{ABCD} \cite{chen2021action}: This dataset contains more than 10K conversations between two human users and has a rich set of diverse conversations spanning 55 user intents in the customer service domain. The key characteristic that makes this dataset relevant for our setting is that the agent is required to follow a set of agent guidelines, which makes the conversations inherently workflow-guided.\vspace{-0.3cm}
\paragraph{MultiWoz} \cite{budzianowski2018multiwoz} This dataset also contains more than 10k dialogues and is commonly used in task-oriented dialogue papers \cite{wu2020tod,budzianowski2019hello,ham2020end}. Prior work \cite{hattami2022workflow} has constructed workflows for this dataset, which we use as the ground truth workflows for training. We found that MultiWoz had less variety in the workflow action space (12 unique actions as compared with 30 for ABCD).

\subsection{Previous metrics}
Inspired by \citet{hattami2022workflow} and \citet{chen2021action}, we evaluate multi-step AST first with exact match accuracy and cascading evaluation.\vspace{-0.2cm}
\paragraph{Exact Match (EM):} This involves computing an exact string match between true and predicted actions. Variants of EM compare the action name (\texttt{action}), slot values (\texttt{value}), or both (\texttt{joint}).\vspace{-0.2cm}
\paragraph{Cascading Evaluation (CE):} Cascading Evaluation gives partial credit to subsequences that are correct (\emph{e.g.} exact match between 3 predicted and 3 true steps, exact match for 2, exact match for 1 and then the scores are averaged). This metric biases towards overgeneration. It is less strict than EM since CE gives partial credit to correct subsequences.

\subsection{Additional metrics}
We also propose several additional metrics to evaluate multi-step AST. BLEU and F1 scores provide different types of partial credit, which can make the comparison of multi-step AST approaches more fine-grained and informative. Graph negative log likelihood measures how well our approaches capture future uncertainty in graph predictions.\vspace{-0.2cm}
\paragraph{Action/Value/Joint BLEU score:} For this metric, we take the top 1 sequence from our graph prediction. We compute BLEU score on the predicted and true sequences. We format the sequence as actions separated by semicolons and include action, value, and joint variants \cite{papineni2002bleu}.\vspace{-0.2cm}
\paragraph{Action/Value/Joint F1-score:} We also compute F1-score, which does not evaluate the ordering of actions but rather precision/recall of an unordered set. It's also evaluated on the top 1 sequence from the graph. This metric is useful for domains where strict ordering of actions is not required.\vspace{-0.2cm}
\paragraph{Action Graph NLL:} Graph negative log-likelihood (NLL) evaluates how well a predicted graph captures uncertainty in future action sequences. It involves traversing the graph and computing NLL of the ground truth sequence given the graph probabilities. This is similar to token-level negative log-likelihood but uses our predicted graph and edge probabilities. As discussed in Section \ref{sec:formulation}, multi-step graphs are important to predict for multi-step AST because they better capture the inherent uncertainty in dialogue that stems from future customer utterances.

\subsection{Models}
For our fine-tuned multi-step AST model, we use \texttt{t5-small} \cite{raffel2020exploring}, as is used in prior work \cite{hattami2022workflow}. For in-context learning models, we set $K=5$ and use \texttt{gpt-3.5-turbo-0301} \cite{brown2020language}. In Appendix~\ref{app:incontext}, we compare to \texttt{gpt-4-0314} \cite{openai2023gpt} with $K=30$ on a subset of the data. We generate $R=20$ rollouts for the fine-tuned models and in-context learning. For graph traversal, we set the edge threshold $t_{edge}$ to 1 and the max prediction sequence length to 7 for ABCD and 11 for MultiWoz, which is the 99th percentile of number of actions on each respective training dataset.

As a baseline, we compare to an approach from prior work that fine-tunes on a 1-step AST task~\cite{hattami2022workflow,chen2021action}. The fine-tuned 1-step AST model, also \texttt{t5-small}, is trained to only predict the next workflow action rather than multiple future actions. At inference time, we generate an action from the model and get a sequence of length 1 from this model. The reason we cannot roll out full action sequences from this model is because we require multi-step prediction with no new context information so we cannot receive new observations of future user or system utterances and make another prediction. Graph traversal is only evaluated on action metrics since we do not represent slot values in the graph.

\begin{table*}[t]
\centering
\small

\begin{tabular}{llrrrrrr}
\toprule
\textbf{Dataset} &      \textbf{Approach} &  \textbf{Action EM} & \textbf{Value EM} & \textbf{Joint EM} &  \textbf{Action CE} & \textbf{Value CE} & \textbf{Joint CE} \\
\midrule
     ABCD &     Fine-tuned 1-step AST &        0.246 &      0.235 &       0.23 &        0.338 &      0.327 &      0.318 \\
      & Fine-tuned multi-step AST &        \textbf{0.501} &      \textbf{0.424} &      \textbf{0.414} &        0.705 &      \textbf{0.624} &      \textbf{0.607} \\
      &       In-context learning &        0.368 &      0.262 &      0.255 &        0.635 &      0.488 &      0.473 \\
      &           Graph traversal &        0.094 &        N/A &        N/A &        \textbf{0.754} &        N/A &        N/A \\
    \midrule
 MultiWoz &     Fine-tuned 1-step AST &        0.174 &      \textbf{0.161} &       \textbf{0.16} &        0.253 &      0.234 &      0.232 \\
  & Fine-tuned multi-step AST &        \textbf{0.177} &      0.103 &      0.102 &        0.531 &      \textbf{0.275} &      \textbf{0.273} \\
  &       In-context learning &        0.140 &      0.029 &      0.029 &        0.419 &      0.092 &      0.091 \\
  &           Graph traversal &        0.006 &        N/A &        N/A &        \textbf{0.840} &        N/A &        N/A \\
\bottomrule
\end{tabular}
\caption{Performance of our approaches on metrics proposed in prior work. The fine-tuned multi-step AST model performs the best on most metrics for ABCD, while graph traversal performs the best on cascading evaluation since this metric prefers overgeneration. The fine-tuned 1-step AST model performs well on MultiWoz as it is sufficient to take greedy actions on this task to do well.}
\label{tab:existing_metrics}
\end{table*}

\begin{table*}[t]
\centering
\small

\begin{tabular}{llrrrrrrr}
\toprule
\textbf{Dataset} & \textbf{Approach} &  \textbf{\thead{Action\\F1 $\uparrow$}} & \textbf{\thead{Value\\F1 $\uparrow$}} & \textbf{\thead{Joint\\F1 $\uparrow$}} &  \textbf{\thead{Action\\BLEU $\uparrow$}} & \textbf{\thead{Value\\BLEU $\uparrow$}} & \textbf{\thead{Joint\\BLEU $\uparrow$}} & \textbf{\thead{Action\\Graph NLL $\downarrow$}} \\
\midrule
     ABCD &     Fine-tuned 1-step AST &        0.537 &      0.553 &      0.501 &         28.465 &       34.283 &        35.53 &              44.106 \\
      & Fine-tuned multi-step AST &        \textbf{0.801} &      \textbf{0.716} &       \textbf{0.69} &         \textbf{66.712} &       \textbf{65.139} &       \textbf{68.945} &              10.371 \\
      &       In-context learning &        0.714 &      0.567 &      0.535 &         55.471 &       50.893 &       55.971 &              19.915 \\
      &           Graph traversal &        0.668 &        N/A &        N/A &         27.960 &          N/A &          N/A &               \textbf{6.342} \\
     \midrule
 MultiWoz &     Fine-tuned 1-step AST &        0.610 &      \textbf{0.405} &      \textbf{0.403} &         24.466 &       23.498 &       24.714 &              48.920 \\
  & Fine-tuned multi-step AST &        0.650 &      0.315 &      0.314 &         \textbf{50.763} &       \textbf{30.989} &       \textbf{43.385} &              22.478 \\
  &       In-context learning &        0.608 &      0.147 &      0.145 &         41.197 &       12.347 &        25.14 &                 38.888${^*}$ \\
  &           Graph traversal &        \textbf{0.745} &        N/A &        N/A &         27.244 &          N/A &          N/A &              \textbf{18.061} \\
\bottomrule
\end{tabular}

\caption{Performance of our approaches on \textit{new} metrics. The fine-tuned multi-step AST model performs the best on most metrics for ABCD, while the 1-step AST model performs well on MultiWoz since actions tend to be inherently more greedy in this setting. Graph traversal performs the best on graph NLL as it captures the future paths more accurately. Due to compute limitations, $^*$ indicates that this value was computed on a subset of 500 test examples.}
\label{tab:f1_bleu_metrics}
\end{table*}

\section{Results}\label{sec:results_analysis}

We now present results on both ABCD and MultiWoz. We evaluate the approaches on prior metrics \cite{hattami2022workflow,chen2021action} as well as our new set of metrics. We also show how multi-step prediction can be used to compute dialogue statistics like task success, and how it can lead to efficiencies with action automation.


\paragraph{Comparison of approaches on metrics proposed in prior work:}
We present results in Table \ref{tab:existing_metrics} comparing the three proposed multi-step prediction approaches to 1-step AST models from previous work on the multi-step prediction task. On ABCD, the fine-tuned multi-step AST model achieves the highest performance on all metrics except cascading evaluation, which graph traversal performs the best on. This is because graph traversal generates longer sequences and CE prefers overgeneration. 

The multi-step action prediction problem is much harder for MultiWoz because conversations are not based on structural workflows like in ABCD. Instead, MultiWoz is better suited for Dialogue State Tracking, where the system needs to identify user intents and slot values. The fine-tuned 1-step AST model does relatively well on this task as greedy action selection is often sufficient. Graph traversal does best on Action CE due to overgeneration and the fine-tuned multi-step AST model achieves the highest value and joint CE scores.

\paragraph{Comparison of approaches on \textit{new} metrics:}
\label{sec:results_new_metrics}

Next, in Table \ref{tab:f1_bleu_metrics}, we compare the performance of all approaches on a new set of metrics. The fine-tuned multi-step AST model outperforms all approaches on F1 and BLEU scores for ABCD and on BLEU scores for MultiWoz. Again, the 1-step AST model does better than the other approaches on MultiWoz because greedy action prediction performs well in this domain. While F1-score mirrors EM for most trends, graph traversal does much better on Action F1 than on Action EM because F1 does not measure the ordering of actions. Since graph traversal does not have access to the context for contextual predictions and relies purely on historical action sequences to predict the future sequence, the order of actions may not always match the ground truth. We've also observed cycles in some of the graphs so predicted action sequences can contain repeated actions, which F1 collapses into one. Thus, F1 is a good measure when the ordering of actions is not a strict requirement and when the sequences contain many repeated actions.

For Action Graph NLL, the fine-tuned 1-step AST model has a very low score because it does not predict a multi-sequence graph. We assign probability $1e-30$ for missing edges so the 1-step model gets heavily penalized for these missing edges. All of our approaches achieve higher performance, with the graph traversal approach performing the best. This is because the full graph constructed from the training data is more comprehensive than graphs constructed from  $R=20$ rollouts of a predictive model. NLL evaluates whether the model's predicted graph accurately models the relationship between actions and the probabilities between them and thus evaluates if the graph is modeling branching accurately. As we explain in Section \ref{sec:formulation}, modeling future branching is important for dialogue multi-step action prediction as it depends highly on future customer responses.

\paragraph{Varying the number of steps predicted in multi-step AST:}
Next, we varied the number of max AST steps $N=[1,...,4]$ predicted by the models during inference time , where $N=1$ is exactly the 1-step AST problem setting. We focus on Action EM and CE metrics as they are used in prior work \cite{hattami2022workflow,chen2021action} and are relevant for the graph traversal approach as well. The fine-tuned multi-step AST model is able to retain performance and even perform slightly better than the 1-step AST model for $N=1$, while doing significantly better as $N$ increases. For Action EM, the multi-step AST model does the best, followed by in-context learning, followed by graph traversal. For Action CE, all three multi-step action prediction models perform similarly while the 1-step model suffers as $N$ increases.

Graph traversal does not do any dynamic prediction based on utterances in the context. Instead, it relies purely on historical counts to estimate the most likely future path. Given that the customer can steer the conversation in many ways, graph traversal performs worse as you get further into the future, but the benefit is that without requiring any training, we can achieve similar performance to the other methods for low values of $N$.

\begin{figure}[t]
    \centering
    \includegraphics[width=0.48\textwidth]{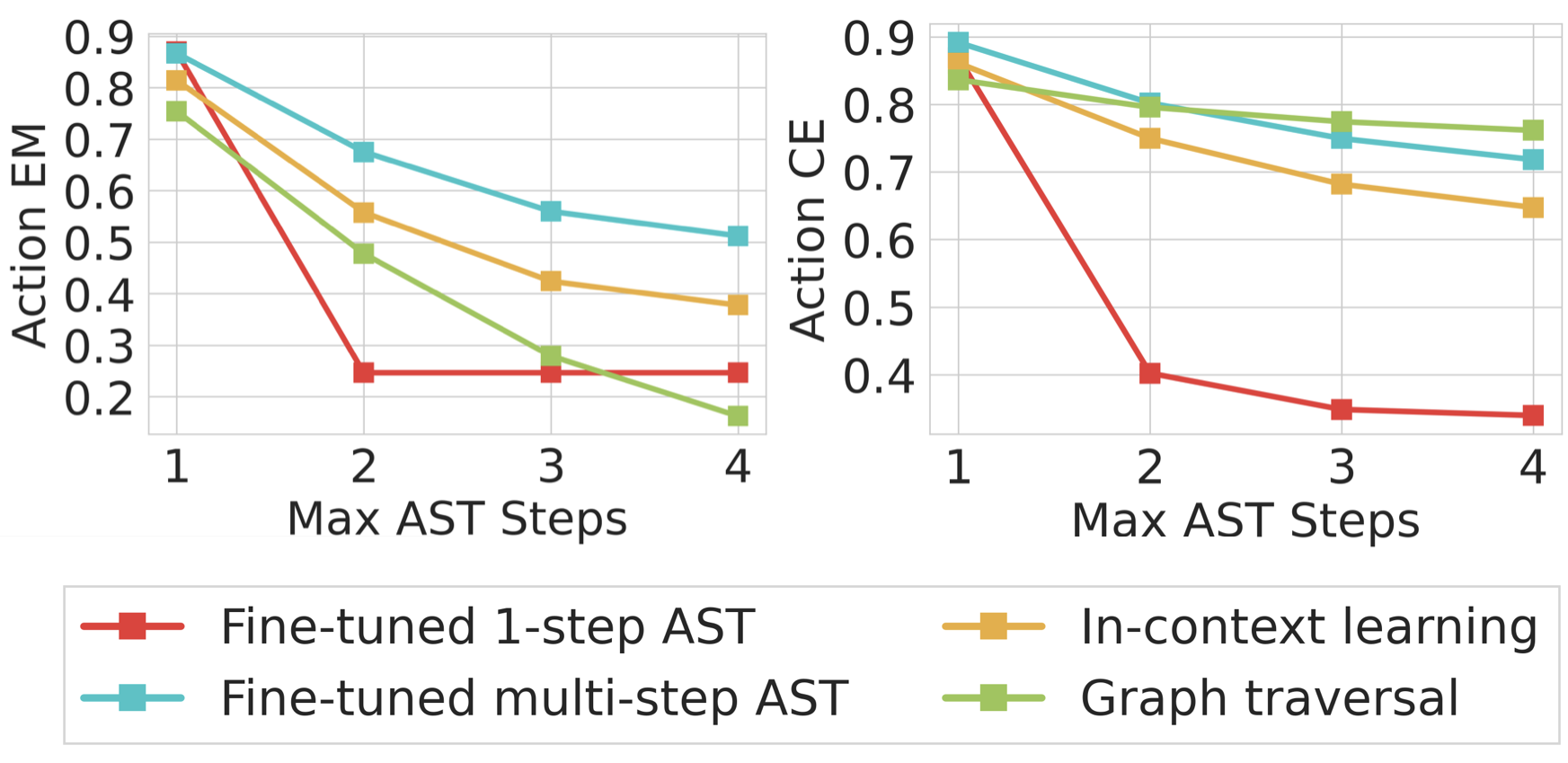}
    \caption{Varying the max number of steps in predicted action sequences on the ABCD dataset. When $N\!=\!1$ (the 1-step AST problem), our multi-step model performs equivalently or better than the 1-step AST model. As we increase $N$, our models perform much better.}
    \label{fig:vary_nstep}
\end{figure}

\section{Applications of multi-step AST}
Next, we use the best method from Section~\ref{sec:results_analysis} (the fine-tuned multi-step AST model) to demonstrate the effectiveness of using multi-step AST on two downstream applications.
\paragraph{Predicting dialogue statistics:}
We show that multi-step actions provide useful information for estimating statistics about a given conversation. Estimating conversation-level statistics has several implications. For example, accurately predicting dialogue success enables the system to escalate certain dialogues to a human for extra intervention. This focuses the human’s attention on where it is needed the most \cite{xu2020ai}. Accurately estimating the number of remaining turns can enable better queueing and scheduling of issues across a call center. This can dramatically decrease long wait times \cite{ostrom2019customer}.

In the middle of the dialogue, we can use a 1-step or multi-step chain of future actions as features to predict quantities that depend on the dialogue's outcome, such as its duration and whether or not it ended in success. Specifically, we fine-tune BERT models~\cite{devlin2019bert} with a regression head (\emph{i.e.} mean-squared loss) to predict different target values. All models include utterances and actions in the dialogue context $C_t$. We improve this representation by adding features for the actions after step $t$: either predictions from the fine-tuned multi-step AST model in Section~\ref{sec:results_analysis} or the oracle ground truth actions for that dialogue. Table~\ref{tab:convo_statistics} shows that concatenating additional actions to $C_t$ leads to improved prediction on the ABCD test set, especially for Dialogue Success and Fraction of Conversation Complete. We also see that predicting all actions in the dialogue is competitive with, and often better than, using a 1-step action oracle.

\begin{table*}[t]
\centering
\small
\begin{tabular}{llrr}
\toprule
                                 \textbf{Task} &            \textbf{Action Features} &  \textbf{MSE $\downarrow$} &  \textbf{Match (rounded) $\uparrow$} \\
\midrule
\multirow{4}{*}{Dialogue success}       &        Predicted 2 Actions &            0.1638 &                      0.7677 \\
                                        &      Predicted All Actions &            0.1532 &                      0.7733 \\
                                        &    1 Action (ground truth) &            0.1614 &                      0.7658 \\
                                        & All Actions (ground truth) &        \bf 0.0465 &                  \bf 0.9448 \\
\midrule
\multirow{4}{*}{Fraction of conversation complete}  &        Predicted 2 Actions &            0.0087 &                      0.8703 \\
                                                    &      Predicted All Actions &            0.0089 &                      0.8672 \\
                                                    &    1 Action (ground truth) &            0.0084 &                      0.8720 \\
                                                    & All Actions (ground truth) &        \bf 0.0048 &                  \bf 0.9035 \\
\midrule
\multirow{4}{*}{Number of remaining utterances} &        Predicted 2 Actions &           16.6416 &                      0.1048 \\
                                                &      Predicted All Actions &           16.4366 &                      0.1308 \\
                                                &    1 Action (ground truth) &           16.1658 &                      0.1344 \\
                                                & All Actions (ground truth) &       \bf 13.3853 &                  \bf 0.1400 \\
\midrule
\multirow{4}{*}{Number of remaining user utterances}    &        Predicted 2 Actions &            5.4087 &                      0.2203 \\
                                                        &      Predicted All Actions &            4.7974 &                  \bf 0.2422 \\
                                                        &    1 Action (ground truth) &            5.0883 &                      0.2353 \\
                                                        & All Actions (ground truth) &        \bf 4.4901 &                      0.2400 \\
\midrule
\multirow{4}{*}{Number of remaining system utterances}  &        Predicted 2 Actions &            4.3595 &                      0.2259 \\
                                                        &      Predicted All Actions &            4.1614 &                      0.2267 \\
                                                        &    1 Action (ground truth) &            4.4609 &                      0.2445 \\
                                                        & All Actions (ground truth) &        \bf 3.5307 &                  \bf 0.2647 \\
\bottomrule
\end{tabular}
\caption{Performance of downstream prediction tasks on the ABCD dataset using different action features. Predicted actions from the fine-tuned model are competitive with and often better than 1 ground truth action.}
\label{tab:convo_statistics}
\end{table*}

\paragraph{Automating dialogue actions:}\label{subsec:simulated_augmentation}


\begin{figure}[t]
    \centering
    \includegraphics[width=0.52\textwidth]{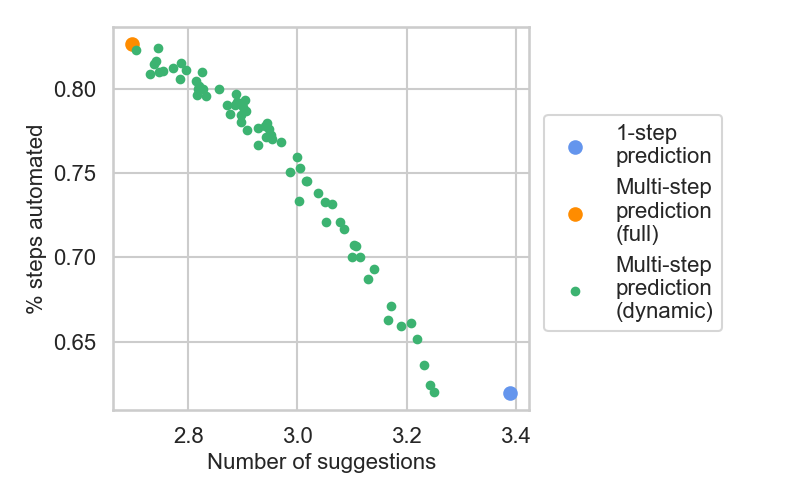}
    \caption{Automation results comparing methods along two axes: \% of steps automated and number of suggestions. Multi-step prediction (full) achieves 20\% more automation of steps compared with 1-step prediction. Multi-step prediction (dynamic) with various model confidence thresholds shows the tradeoff of more automation vs. more human involvement.}
    \label{fig:automation_simulation_results}
\end{figure}

In this section, we show that predicting multi-step actions can help agents automate the conversation, leading to greater efficiency. In this experiment, we assume that the agent has access to a large language model that can take over the conversation with 100\% success, without getting derailed by what the customer says. However, it can only automate parts of the conversation when the next action predictions are ``approved'' as accurate. We compare three agent assistance modes using the ABCD dataset. 

First, we look at 1-step prediction, where we simulate showing the next workflow action to the agent upon completion of the previous workflow action or at the beginning of the conversation. Then, we decide that the agent approved the workflow action if it matches the ground truth next workflow action. We make at most $N$ model calls for $N$ workflow actions within a conversation, and calculate the results based on the number of times the model got the prediction correct. In this scenario, if the model gets all of the predictions correct, then the model would have been called $N$ number of times for a conversation with $N$ workflow actions.  

Next, we look at multi-step prediction (full), where we simulate showing the full predicted workflow sequence at once to the agent at the beginning of the conversation. We simulate the agent approving the correct actions by comparing with the ground truth. Once the approved actions are performed, we prompt the model again to predict the rest of the workflow actions. We continue this until the model gets all of the actions correct or until the number of model calls equal the number of workflow actions needed to be performed. We finally consider multi-step prediction (dynamic), which leverages model confidence scores to determine the ideal length of the predicted sequence. To do this, we use the model's transition scores and select various thresholds to determine the ideal point to stop prediction. For all methods, we calculate:\vspace{-0.3cm}
\paragraph{1) \% steps automated:} Average number of workflow actions successfully automated as a percentage of all actions in the conversation. Higher is better since this directly indicates the percentage of actions the model was able to correctly predict.\vspace{-0.3cm}
\paragraph{2) Number of suggestions:} Average number of suggestions shown to the agent. Lower is better given prediction cost, context switching for the agent when reading and approving, etc.





The results in Figure~\ref{fig:automation_simulation_results} show that multi-step prediction (full) is able to automate an additional 20\% of workflow actions, compared to 1-step prediction. However, there are fewer suggestions displayed to an agent, which means less human involvement. 1-step prediction, on the other hand, has much less automation, but we would provide a suggestion at every action step. Multi-step prediction (dynamic) uses model confidence scores to determine how much of the sequence to predict, which allows a system designer to decide how much human interaction should be required. We can achieve any point on this tradeoff between \% automation and human involvement based on the application.





\section{Conclusion}
In this paper, we propose the novel problem of multi-step Action State Tracking for task-oriented dialogue. Predicting multiple workflow actions can be useful for automating actions and increasing efficiency of conversations. We develop three approaches: 1) fine-tuning a model on multi-step action sequences, 2) using few-shot in-context learning with large language models, and 3) constructing graphs from historical workflows to use for prediction. Results show that while the fine-tuned multi-step model performs the best on most metrics, it requires a full training procedure. The in-context learning approach with a powerful LLM like GPT-4 is competitive with the fine-tuned multi-step model but is costly. Finally, the graph traversal approach is able to capture much of the inherent uncertainty in future action sequences through branching. 

\section{Limitations}
Our AST approaches assume access to annotated, aligned actions for each dialogue. Since these may not be available in general, one interesting extension of this formulation would be to begin with unlabeled dialogues and induce an action schema and action workflows.

Generating multi-step graphs are more costly than single-step generation in terms of both computation and latency. Moreover, suggesting multi-step action predictions to human agents in practice may lead to additional cognitive load. Optimizing the user experience is an important design problem beyond the scope of this work. When simulating partial automation, our setup does not account for the complex relationship between accepting suggestions, automating actions, automating utterances, and completing dialogues more quickly.

\bibliography{references}

\appendix
\section{Experiment details}
Table \ref{tab:experiment_parameters} includes the hyperparameters used for the fine-tuning approach (Section \ref{sec:fine_tuning}).

\begin{table}[!ht]
\centering
\small
\begin{tabular}{l||c}
\toprule
\multicolumn{2}{c}{Fine-tuning approach} \\
\midrule
Model & \texttt{t5-small} \\
Training epochs & 30\\
Learning Rate & 5e-5 \\
Max source length & 1024 \\
Max target length & 256 \\
Warm-up steps & 500 \\
\bottomrule
\end{tabular}
\caption{Experiment Models \& Parameters}
\label{tab:experiment_parameters}
\end{table}
\section{In-context learning ablation}\label{app:incontext}
Table~\ref{tab:incontext_prediction_ast_results} shows how different design parameters affect performance of the in-context learning approach on a subset of the ABCD test dataset. First, we note that GPT-4 outperforms the corresponding GPT-3.5 model by as much as 6\%. GPT-4 on this subset of the data also outperforms the fine-tuned multi-step AST model on the full dataset in terms of cascading evaluation metrics. Removing action contexts from the context retrieval features leads to a slight drop in exact match accuracy and a slight increase in cascading evaluation metrics. Other feature combinations lead to much worse performance.

\begin{table*}[ht!]
\centering
\small
\begin{tabular}{lllrrrrrr}
\toprule
          \textbf{Model} & \textbf{\thead{Retrieval\\Features}} &          \textbf{\thead{Prediction\\Features}} &  \textbf{\thead{Action\\EM}} &  \textbf{\thead{Value\\EM}} &  \textbf{\thead{Joint\\EM}} &  \textbf{\thead{Action\\CE}} &  \textbf{\thead{Value\\CE}} &  \textbf{\thead{Joint\\CE}} \\
\midrule
\multirow{6}{*}{gpt-3.5, $K\!=\!5$} & Utterance/Action Ctx & Utterance/Action Ctx &      \bf 0.380 &      \bf 0.280 &      \bf 0.270 &     0.6288 &    0.4961 &    0.4771 \\
& Utterance/Action Ctx &              Utterance Context &       0.340 &      0.276 &      0.264 &     \bf 0.6493 &    \bf 0.5394 &    \bf 0.5180 \\
             & Action Context & Utterance/Action Ctx &       0.206 &      0.156 &      0.134 &     0.3751 &    0.2942 &    0.2365 \\
             & Action Context &              Utterance Context &       0.196 &      0.128 &      0.114 &     0.3828 &    0.2802 &    0.2337 \\
& Utterance/Action Ctx &              Action Context &       0.142 &      0.064 &      0.048 &     0.4543 &    0.2351 &    0.1981 \\
             & Action Context &              Action Context &       0.066 &      0.062 &      0.034 &     0.1390 &    0.1518 &    0.0709 \\
\midrule
\multirow{6}{*}{gpt-4, $K\!=\!30$} & Utterance/Action Ctx & Utterance/Action Ctx &       \bf 0.446 &      \bf 0.358 &     \bf 0.346 &     0.7362 &    0.6133 &    0.5969 \\
& Utterance/Action Ctx &              Utterance Context &       0.394 &      0.330 &      0.314 &     \bf 0.7491 &    \bf 0.6566 &    \bf 0.6368 \\
             & Action Context & Utterance/Action Ctx &       0.292 &      0.210 &      0.190 &     0.5956 &    0.4842 &    0.4468 \\
             & Action Context &              Utterance Context &       0.286 &      0.220 &      0.202 &     0.5865 &    0.4808 &    0.4475 \\
& Utterance/Action Ctx &              Action Context &       0.204 &      0.120 &      0.118 &     0.7050 &    0.4516 &    0.4356 \\
             & Action Context &              Action Context &       0.102 &      0.064 &      0.064 &     0.3804 &    0.2283 &    0.2014 \\
\bottomrule
\end{tabular}
\caption{In-context results using 500 samples from the ABCD test set for different language models, retrieval features, and prediction features. GPT-4 is more accurate than GPT-3.5, and removing action context from the prediction model results in slightly lower accuracy results and slightly higher cascading evaluation results.}
\label{tab:incontext_prediction_ast_results}
\end{table*}

\end{document}